\documentclass{article}

\usepackage{PRIMEarxiv}

\usepackage[utf8]{inputenc} % allow utf-8 input
\usepackage[T1]{fontenc}    % use 8-bit T1 fonts
\usepackage{hyperref}       % hyperlinks
\usepackage{url}            % simple URL typesetting
\usepackage{booktabs}       % professional-quality tables
\usepackage{amsfonts}       % blackboard math symbols
\usepackage{nicefrac}       % compact symbols for 1/2, etc.
\usepackage{microtype}      % microtypography
\usepackage{lipsum}
\usepackage{fancyhdr}       % header
\usepackage{graphicx}       % graphics
\graphicspath{{media/}}     % organize your images and other figures under media/ folder
\usepackage{amssymb}
\usepackage{float}
\usepackage{subfigure}
\usepackage{amsmath}
\usepackage{algorithm}
\usepackage{algpseudocode}
\usepackage{threeparttable}
\usepackage{array}
\usepackage{multirow}
\usepackage{color,xcolor}
\usepackage{epstopdf}
\usepackage{bm}
\usepackage{amsopn}
\usepackage{amssymb,amsmath,color,times}
\usepackage{lineno}

\newcommand{\etal}{\textit{et al.}}
\title{DMNR: Unsupervised De-noising of Point Clouds Corrupted by Airborne Particles
}

\author{
  Chu Chen \\
  Department of Mathematics \\
  City University of Hong Kong \\
  Hong Kong\\
  \texttt{{chuchen4-c}@my.cityu.edu.hk} \\
    \And
  Yanqi Ma, Bingcheng Dong\\
  School of Mathematical Science \\
  Dalian University of Technology \\
  Dalian\\
  \texttt{{mayq697, 18352203288}@mail.dlut.edu.cn} \\
   \AND
   Junjie Cao \thanks{Corresponding Author.}\\
   School of Mathematical Science \\
   Dalian University of Technology \\
   Dalian\\
   \texttt{jjcao@dlut.edu.cn} \\
}

\begin{document}
\maketitle

\begin{abstract}
LiDAR sensors are critical for autonomous driving and robotics applications due to their ability to provide accurate range measurements and their robustness to lighting conditions. However, airborne particles, such as fog, rain, snow, and dust, will degrade its performance and it is inevitable to encounter these inclement environmental conditions outdoors. It would be a straightforward approach to remove them by supervised semantic segmentation. But annotating these particles point wisely is too laborious. To address this problem and enhance the perception under inclement conditions, we develop two dynamic filtering methods called Dynamic Multi-threshold Noise Removal (DMNR) and DMNR-H by accurate analysis of the position distribution and intensity characteristics of noisy points and clean points on publicly available WADS and DENSE datasets. Both DMNR and DMNR-H outperform state-of-the-art unsupervised methods by a significant margin on the two datasets and are slightly better than supervised deep learning-based methods. Furthermore, our methods are more robust to different LiDAR sensors and airborne particles, such as snow and fog.
\end{abstract}

% keywords can be removed

\section{Introduction}
As a commonly used 3D measurement sensor for machine environmental perception, LiDAR has been widely used in automatic driving, scene reconstruction, and so on. By directly obtaining the exact distance and azimuth of the surrounding objects, LiDAR has the natural advantage of high precision compared to other sensors. However, when it comes to outdoor application scenarios like automatic driving, the scanned point clouds are vulnerable to airborne particles such as snowflakes (Fig.~\ref{intro_rawsnow}), raindrops, fog (Fig.~\ref{intro_rawfog}), dust, and so on. A large number of noise points will be generated which have serious adverse effects on downstream tasks~\cite{Bos2020AutonomyAT}.

To remove the corruption of airborne particles, some efforts have been made. They are mainly divided into conventional filter-based methods and deep learning-based approaches. Deep learning perception methods generally tend to achieve better quantitative results~\cite{Heinzler_2020, Sandler2018MobileNetV2IR}.
However, it is not a proper option to deploy these pre-processing networks together with their downstream applications on one edge device with limited computational resources.

Besides, the dependency on point-wise annotations also limits their application. Although there are some efficient 3D point annotation methods, manual annotation of 3D airborne particles is still laborious and time-consuming. Because it is hard to take advantage of the assistance of camera view when labeling these particles, since they may be not visible in camera images or the images are also inconsistently degraded~\cite{Nayar1999VisionIB, Khan2019SnowDU, Mao2014DetectingFI}. 
In contrast, the conventional filter-based methods not only have the advantage in this aspect, but they also have no training cost and significantly fewer computational resources.

The conventional filter-based methods~\cite{Charron2018DenoisingOL, Kurup2021DSORAS, Park2020FastAA, Wang2022ASA} can filter out noise points effectively. Meanwhile, a large number of environmental details are misjudged, as shown in Fig.~\ref{intro_dsorsnow}-\ref{intro_ddiorsnow} and Fig.~\ref{intro_dsorfog}-\ref{intro_ddiorfog}. 
More environment points can also be preserved by lowering the aggressiveness of the methods but with the cost of unsatisfying filtering results. 
Since neither of the two situations are acceptable to downstream algorithms, the ability to make a balance is rather crucial for classical filters. What's more, previous methods fail to maintain the de-noising effect when dealing with different kinds of airborne particles.

The paper addresses these issues by proposing an unsupervised airborne particle removal strategy for LiDAR in adverse weather. 
By staging multiple filtering processes with adaptive spatial and intensity thresholds, our strategy shows its effective de-noising performance and robustness when dealing with various airborne particles and LiDAR sensors (32 or 64 channels). 
Following the strategy, we introduce two algorithms called Dynamic Multi-threshold Noise Removal (DMNR), which is a standard multi-stage filtering method with multiple dynamic thresholds, and DMNR-H, which includes an extra post-processing step by HDBSCAN. The contributions of this paper can be described as follows:

\begin{figure*}[!ht]
    \centering
    \subfigure[Snowy Condition]{
        \label{intro_rawsnow}
        \includegraphics[width=0.22\linewidth]{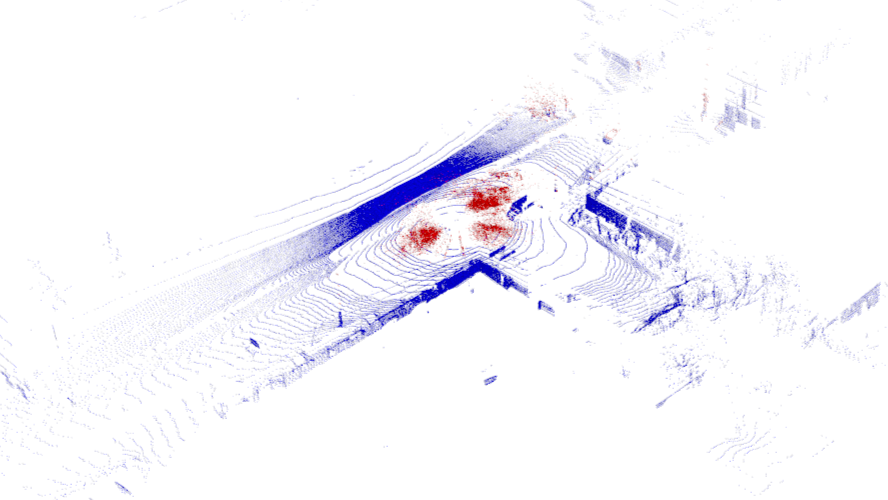}}
    \subfigure[DSOR]{
        \label{intro_dsorsnow}
        \includegraphics[width=0.22\linewidth]{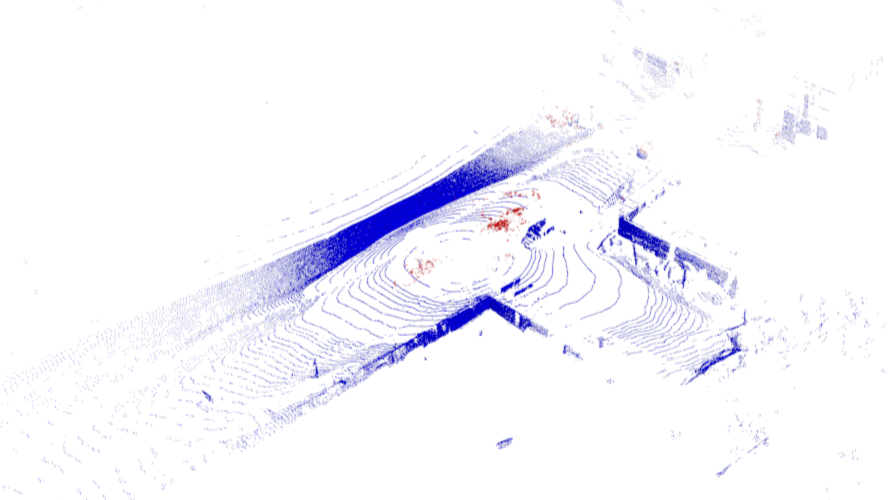}}
    \subfigure[DDIOR]{
        \label{intro_ddiorsnow}
        \includegraphics[width=0.22\linewidth]{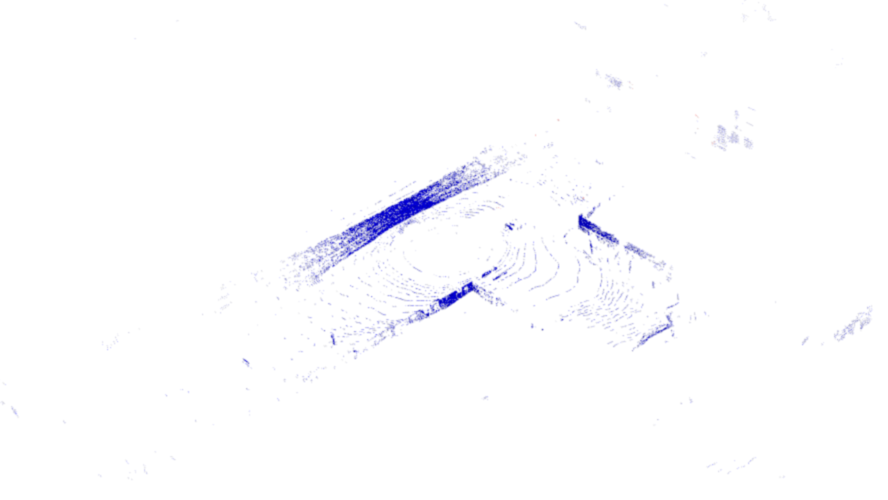}}
    \subfigure[DMNR]{
        \label{intro_dmorsnow}
        \includegraphics[width=0.22\linewidth]{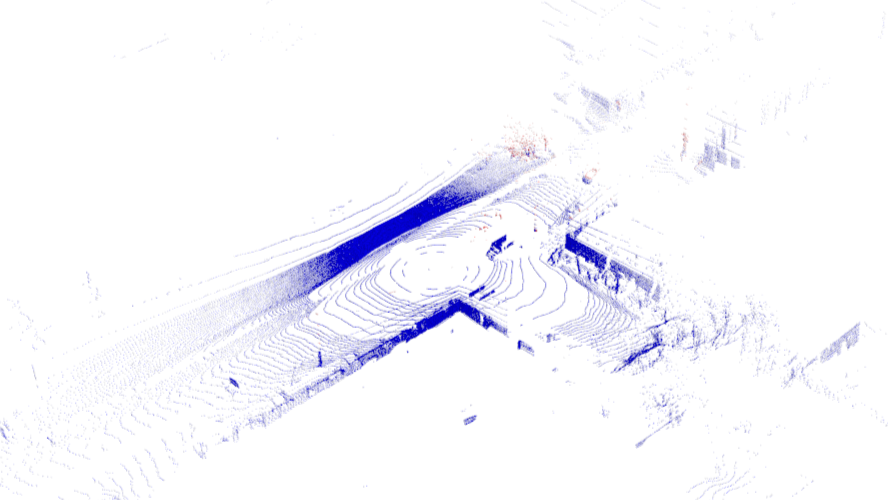}}
    
    \subfigure[Foggy Condition]{
        \label{intro_rawfog}
        \includegraphics[width=0.22\linewidth]{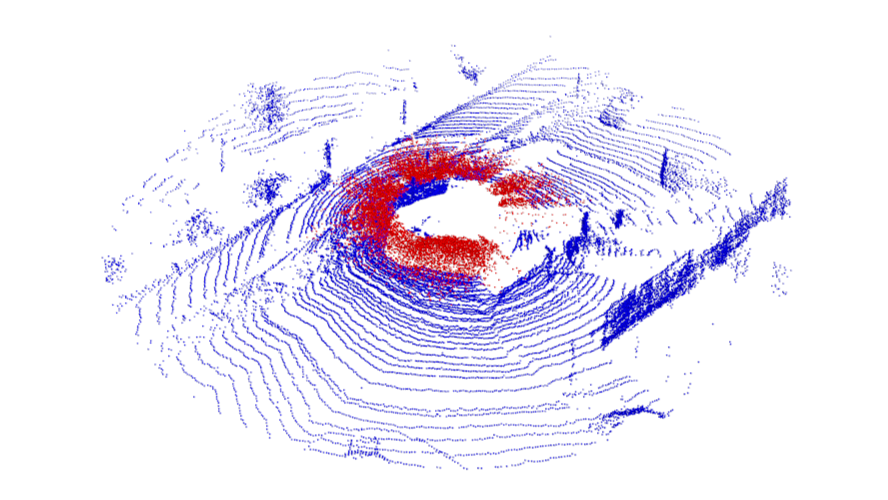}}
    \subfigure[DSOR]{
        \label{intro_dsorfog}
        \includegraphics[width=0.22\linewidth]{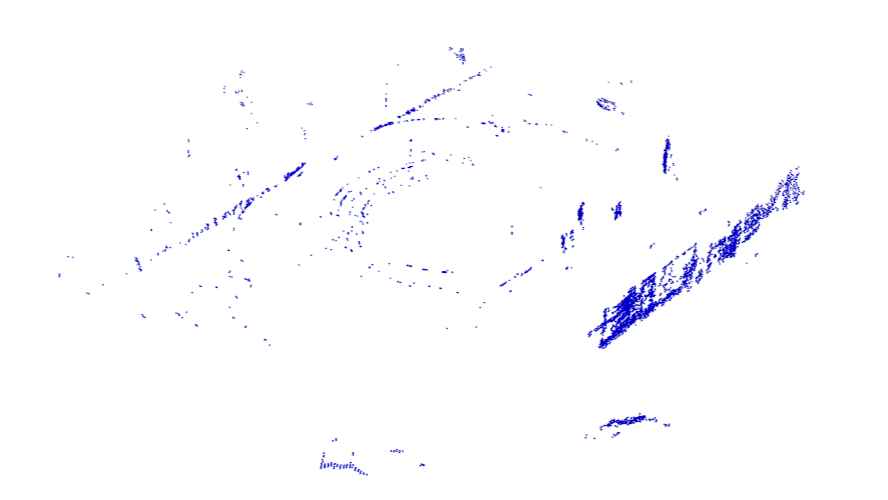}}
    \subfigure[DDIOR]{
        \label{intro_ddiorfog}
        \includegraphics[width=0.22\linewidth]{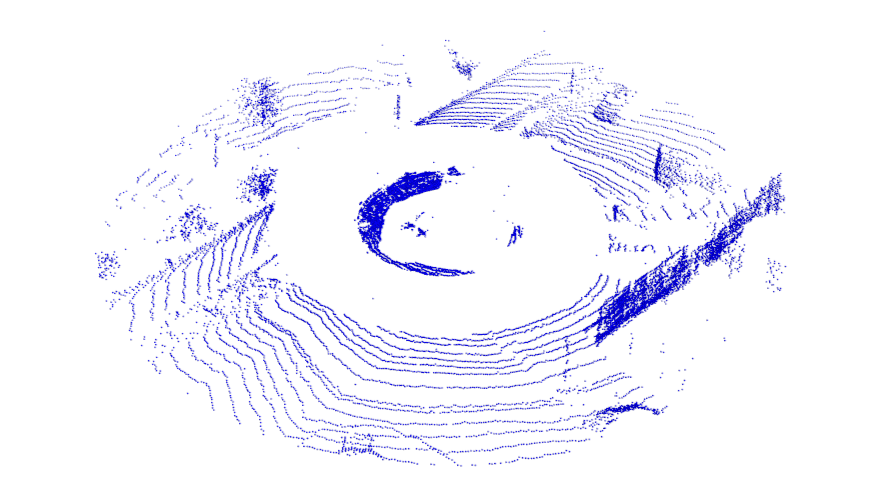}}
    \subfigure[DMNR]{
        \label{intro_dmorsfog}
        \includegraphics[width=0.22\linewidth]{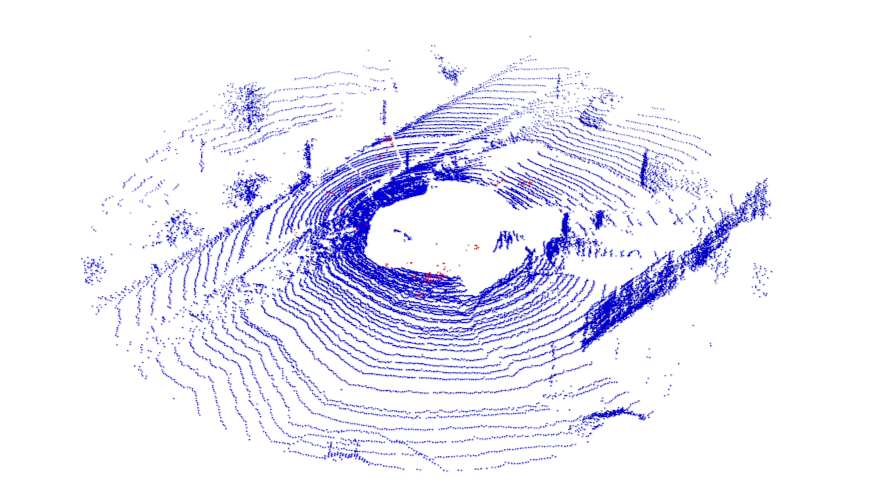}}
    \caption{The proposed DMNR method outperforms previous unsupervised methods for snow and dog removal. The left column shows a snowy point cloud and a foggy point cloud. The rest columns are results of DSOR~\cite{Kurup2021DSORAS}, DDIOR~\cite{Wang2022ASA}, and our DMNR. Noise points are rendered in \textcolor{red}{red} and non-noise points are in \textcolor{blue}{blue}.}
    \label{fig:intro}
\end{figure*}

\begin{itemize}
    \item Two unsupervised LiDAR point cloud de-noising methods are proposed. They follow a multi-stage filtering strategy with multiple dynamic thresholds determined by a thorough analysis of the position distribution and intensity characteristics of the scanned points on the WADS and DENSE datasets.
    % spatial-feature-conformed thresholds
    \item Compared with previous unsupervised methods, our methods are more robust to different kinds of airborne particles and LiDAR devices. 
    \item Quantitative and qualitative evaluations show that our methods outperform previous methods by a large margin for airborne particle removal.
\end{itemize}

\section{Related Works}
LiDAR sensors are vulnerable to airborne particles caused by adverse weather conditions such as rain, snow, and fog. 
In these conditions, airborne particles may result in  intensity attenuation and spurious coordinates of scanned points. Therefore, noise filters are crucial to the quality of the point cloud data. Some denoising algorithms dealing with these noises have been developed in recent years. We would like to focus on unsupervised methods. 

The Point Cloud Library (PCL)~\cite{Rusu20113DIH} contains several filters such as Statistical Outlier Removal (SOR) and Radius Outlier Removal (ROR). But none of these filters are specifically designed for noisy point clouds corrupted by airborne particles, and their performance is far from ideal. 
To deal with point clouds affected by snowflakes, Charron \etal~\cite{Charron2018DenoisingOL} proposed Dynamic Radius Outlier Removal (DROR) by improving ROR to a dynamic search radius in order to preserve more environmental details far from LiDAR sensor. Kurup \etal~\cite{Kurup2021DSORAS} also proposed a de-snowing algorithm called Dynamic Statistical Outlier Removal (DSOR). DSOR can remove more snow than DROR, but it also removes more environmental features. 
%Their ability to preserve environmental details of both the two methods is not ideal, which is not conducive to the subsequent tasks.

Park \etal~\cite{Park2020FastAA} noticed that the intensity of snow points is generally lower than that of non-snow points. Based on this observation, they proposed Low-intensity Outlier Removal (LIOR). However, under heavy snow conditions, when the intensity of snow points is close to that of other points, the performance of LIOR will be reduced significantly. The latest de-snowing algorithm is the Dynamic Distance-intensity Outlier Removal (DDIOR) proposed by Wang \etal~\cite{Wang2022ASA}. DDIOR has a pre-processing step to retain points with high intensity and long distances and adds intensity information into the dynamic threshold. Compared with DSOR, DDIOR slightly improves the precision of snow removal, but it still cannot fully retain environmental details. It still needs to modify parameters to maintain performance when handling different scenes.

There is no unsupervised algorithm that can effectively remove the noise caused by multiple kinds of airborne particles at present. In this work, we compare our DMNR and DMNR-H filters with DSOR and DDIOR on de-snowing based on Winter Adverse Driving dataSet (WADS~\cite{Kurup2021DSORAS}). In addition, we also show the superiority of our methods for different airborne particles by applying snow simulation~\cite{Hahner2022LiDARSS} and fog simulation~\cite{Hahner2021FogSO} on DENSE dataset~\cite{Bijelic_2020_STF}. Note that the LiDAR devices used in the two datasets are also different. 

\section{Methods}
\label{sec:method}
\subsection{DMNR}
\begin{figure}[!ht]
    \centering
    \includegraphics[width=83mm]{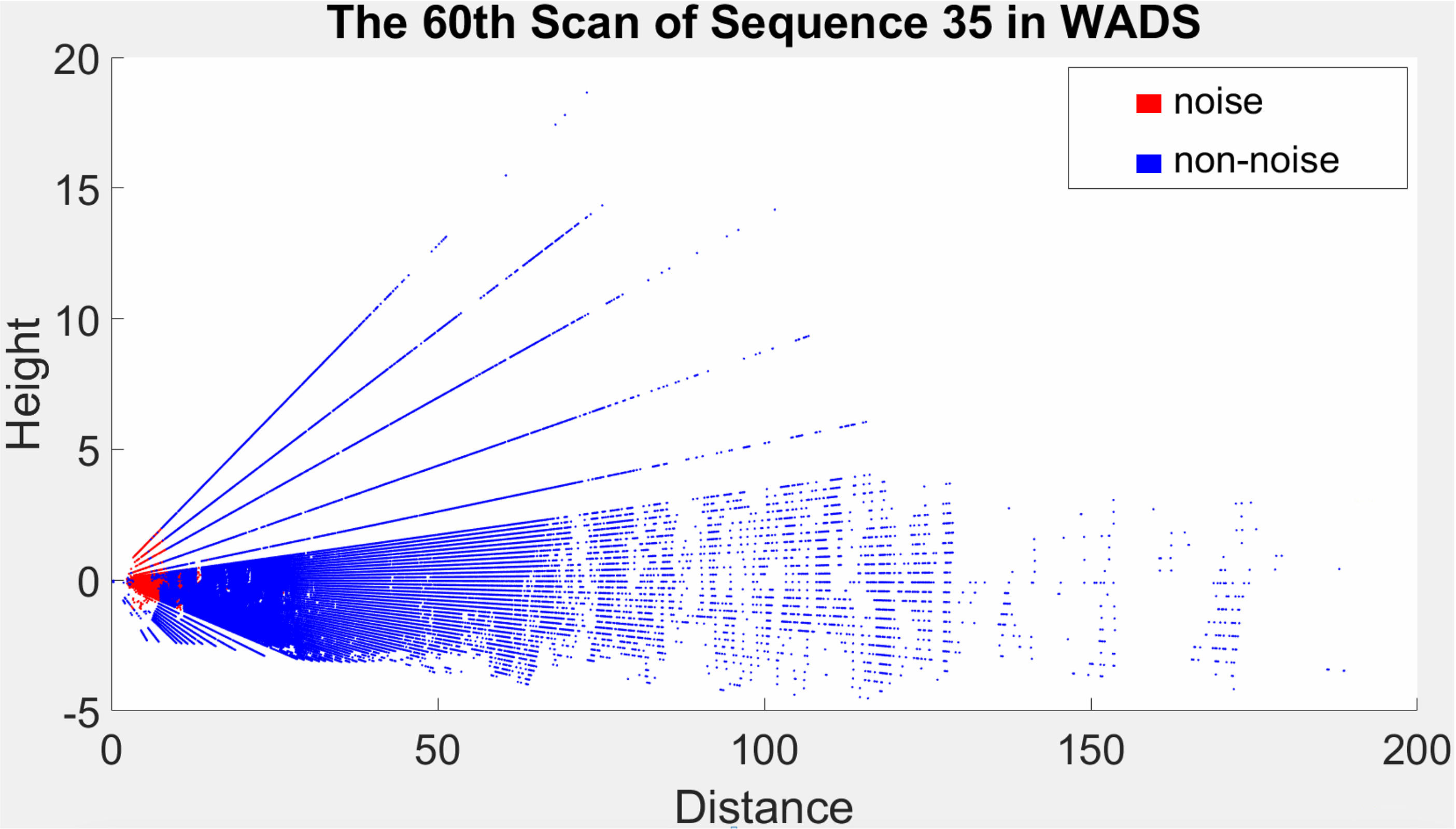}
    \caption{Relationship between distance from sensor and height ($z$ value) of scanned points. Noise points are rendered in \textcolor{red}{red}, and non-noise points in \textcolor{blue}{blue}.}
    \label{fig:H-D relation}
\end{figure}

\begin{figure}[!ht]
    \centering
    \subfigure[Raw Point Cloud]{
        \label{Raw Hor}
        \includegraphics[width=65mm]{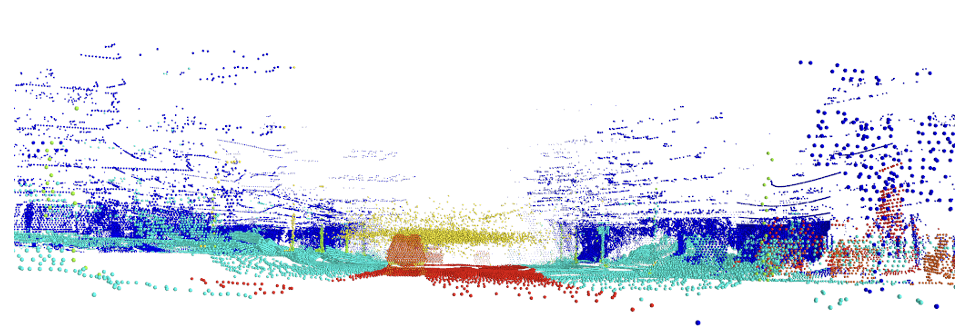}}
    \subfigure[DSOR]{
        \label{DSOR Hor}
        \includegraphics[width=65mm]{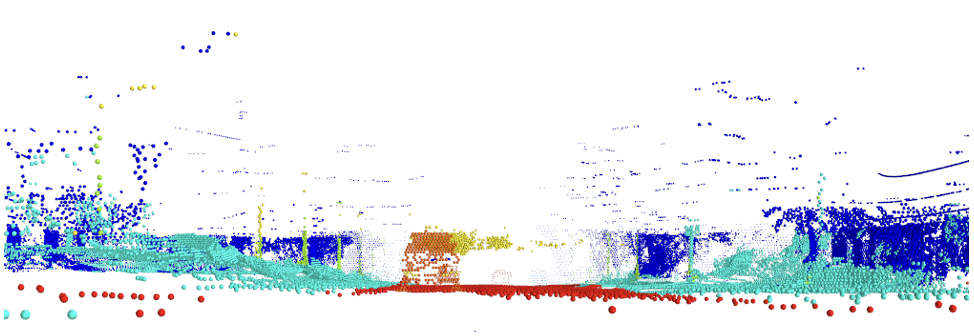}}
    \subfigure[DDIOR]{
        \label{DDIOR Hor}
        \includegraphics[width=65mm]{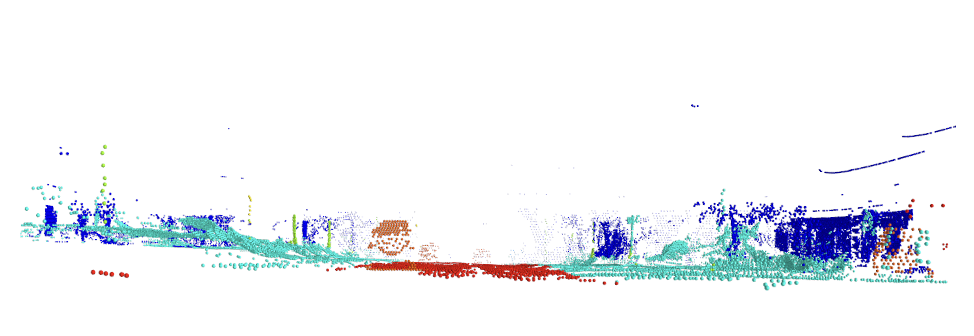}}
    \subfigure[DMNR]{
        \label{DMNR Hor}
        \includegraphics[width=65mm]{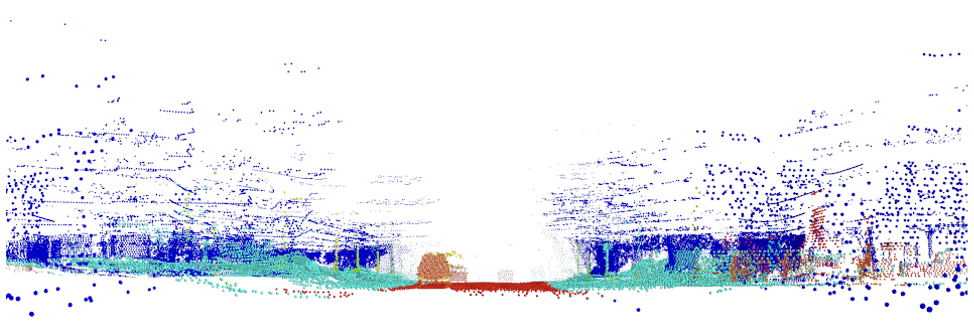}}
    \caption{Horizontal view of a raw point cloud (a) and the results of DSOR (b), DDIOR (c), and DMNR (d). The point cloud are rendered according to their labels and noise points are rendered in \textcolor{yellow}{yellow}.}
    \label{fig:preprocessing function}
\end{figure}

Wang \etal\cite{Wang2022ASA} analyzed the fundamental characteristics of snow noise points including low density, low intensity, close range, and fast decay. Hahner \etal~\cite{Hahner2022LiDARSS} also reached a similar conclusion about fog noise from an optical perspective. These characteristics of these noise points are very important to the study of de-noising algorithms. But they all ignored the height characteristics of the noisy points. Fig.~\ref{fig:H-D relation} shows the height of the points in different ranges. 
Clean points collected at higher altitudes tend to have lower densities. 
It can be seen from Fig.~\ref{fig:preprocessing function} that previous de-noising algorithms failed to distinguish clean points at high places from noise points, resulting in misjudgments and loss of environmental details.

The first stage of DMNR is to distinguish clean environmental points from noise points by a height threshold. 
For each point $p_n=(x_n,y_n,z_n)$, we have:
\begin{equation}
  H_n = \frac {h_1} {d_n}+h_2 ,
  \label{eq:height threshold}
\end{equation}
where $d_n=\sqrt{x_n^2+y_n^2+z_n^2}$ is the distance from the point to the sensor. When dealing with one scan of point cloud, $h_1$ and $h_2$ are constants, which can be determined by $h_1=\mathop{max}\limits_{i=1,...,N} \left\{d_i/2\right\}$ and $h_2=\mathop{min}\limits_{i=1,...,N} \left\{z_i-1\right\}$ respectively. Points with higher height ($z_n>H_n$) will be retained directly, while those with lower height ($z_n\leq H_n$) will go to the next stage of the algorithm.

The second stage of DMNR is to filter out noise points by a point-wise dynamic threshold about density, intensity, and range:
\begin{equation}
 T_n = \mu\times(k_1\times {\rm e}^ {k_2 \times d_n} + k_3 \times i) \times d_n, 
  \label{eq:dynamic threshold}
\end{equation}
where $k_1$ is an indirect aggressive parameter, $k_2$ is an aggressive range parameter, $i$ is the intensity of the point, $k_3$ is the weight of intensity, and $\mu$ is the global average density.
$\mu$ measures the overall density of the point cloud. It can be obtained by averaging the local average density $ad_n$:
\begin{equation}
 \mu=\frac{\sum_{n=1}^N ad_n}{N}, 
  \label{eq:global average distance}
\end{equation}
where $N$ is the number of points.
$ad_n$ is computed as the mean distance of point $p_n$ and its K nearest neighbors.

We can control the aggressive level of DMNR through $k_1$ and $k_2$, which can be determined according to weather conditions and the range of noise points respectively. Algorithm 1 shows the pseudo-code of DMNR. DMNR divides the raw point cloud $P$ into two sets: the filtered point cloud $F$ and the outliers $O$.

\begin{algorithm}[!ht]
    \caption{Dynamic Multi-threshold Outlier Removal}
    \label{algDMOR}
    \begin{algorithmic}[1]
        \Require
            Point Cloud $P=p_n,p_n=(x_n,y_n,z_n,i_n),n=1,...,N$\newline
            $K=$number of nearest neighbors\newline
            $k_1=$aggressive parameter\newline
            $k_2=$aggressive range parameter\newline
            $k_3=$weight of intensity
        \Ensure
            Filtered Point Cloud $F=p_{n_j},p_{n_j}\in P,j=1,...,N'$\newline
            Outliers $O=P\backslash F$
            \State $P\leftarrow KdTree$;
 
        \For {$p_n\in P$}
            \State $average\_distances$ $ad_n\leftarrow nearestKsearch(k)$;
        \EndFor\newline
        calculate:global average distance $\mu\leftarrow ad_n$
        \For {$p_n\in P$}
            \State $distance d_n\leftarrow\sqrt{x_n^2+y_n^2+z_n^2}$\newline
            \State $H_n\leftarrow 100/d_n-5$
            \If {$z_n > H_n$}
                \State $F\leftarrow p_n$
            \Else
                \State $T_n\leftarrow\mu\times(k_1\times {\rm e}^ {k_2 \times d_n} + k_3 \times i_n) \times d_n$
                \If {$ad_n < T_n$}
                    \State $F\leftarrow p_n$
                \Else
                    \State $O\leftarrow p_n$
                \EndIf
            \EndIf
        \EndFor \\
        \Return $F,O$
    \end{algorithmic}
\end{algorithm}

\subsection{DMNR-H}
\begin{figure}[!ht]
    \centering
    \includegraphics[width=130mm]{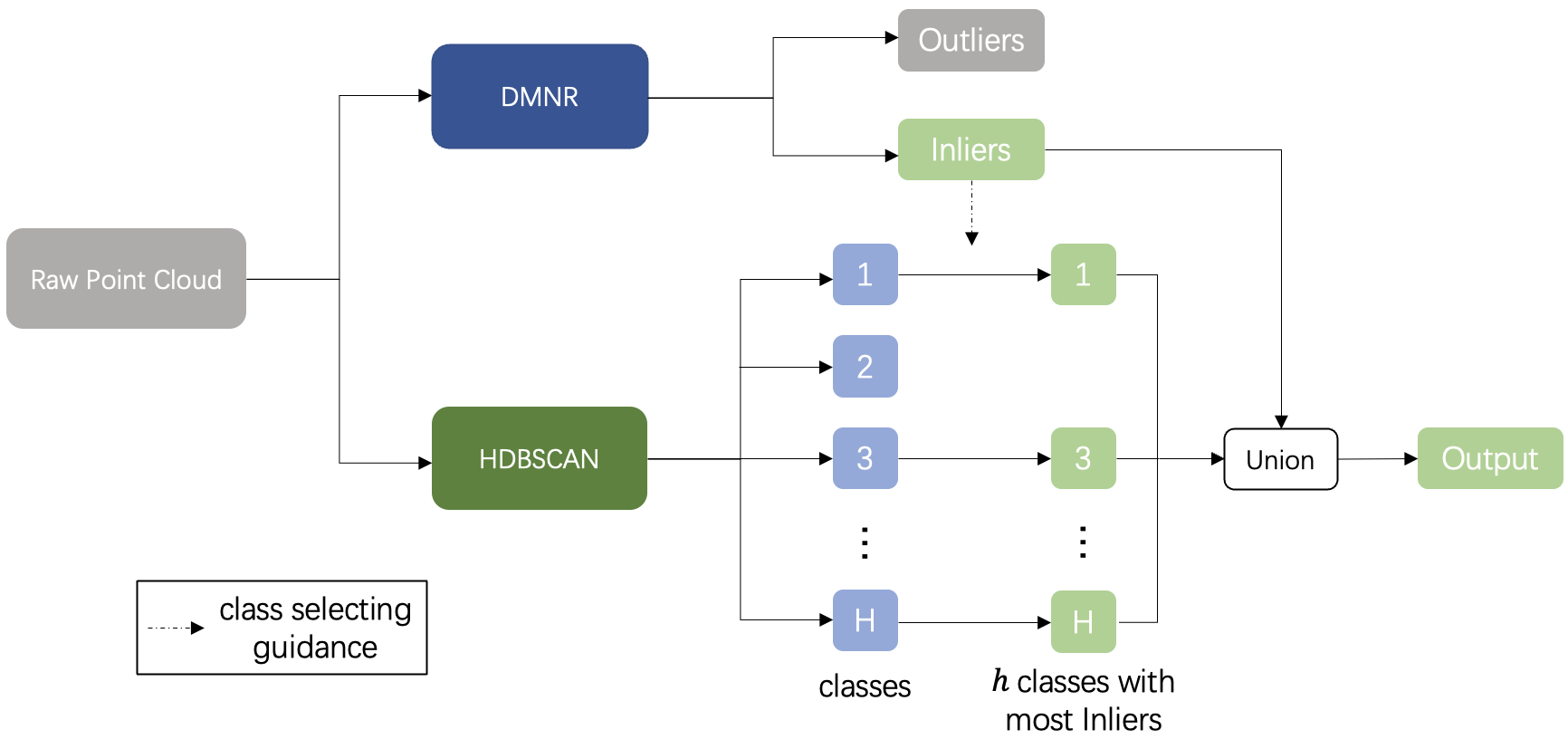}
    \caption{Process of DMNR-H algorithm.}
    \label{fig:DMNRH process}
\end{figure}

Hierarchical Density-Based Spatial Clustering of Applications with Noise (HDBSCAN)~\cite{Campello2013DensityBasedCB} is a clustering algorithm, which can process data with uneven density and large differences in cluster spacing, which are the significant characteristics of point cloud data. 
We use it as a post-processing method to restore some misclassified clean points from the noise points $O$ detected by DMNR.
The point cloud is clustered into multiple classes by HDBSCAN adaptively. DMNR-H will select $h$ classes containing most of the filtered point cloud $F$, and the rest points of the same $h$ classes will be transferred from $O$ to $F$ as clean points. Complete process of DMNR-H can be referred to Fig.~\ref{fig:DMNRH process}.
Consequently the results of DMNR are effectively improved and the environmental structure can be better preserved.

\section{Experiments}
\label{sec:exp}
\begin{figure*}[!ht]
    \centering
    \subfigure[Raw Point Cloud]{
        \label{Raw}
        \includegraphics[width=0.42\linewidth]{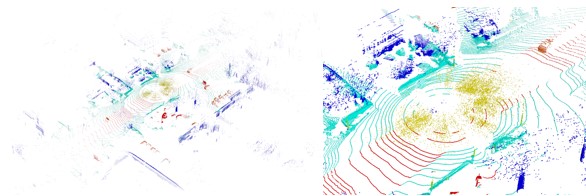}}
    \subfigure[DSOR]{
        \label{DSOR}
        \includegraphics[width=0.48\linewidth]{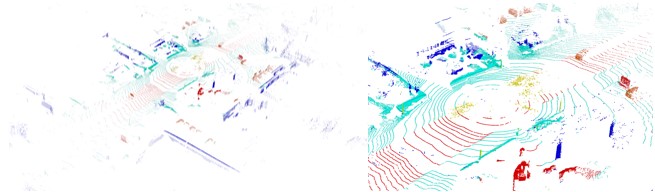}}
    \subfigure[DDIOR]{
        \label{DDIOR}
        \includegraphics[width=0.45\linewidth]{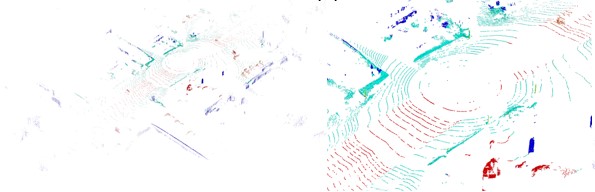}}
    \subfigure[DMNR]{
        \label{DMNR}
        \includegraphics[width=0.48\linewidth]{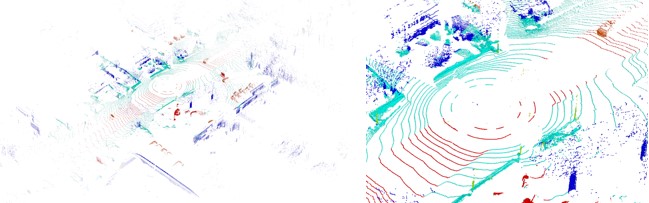}}
    \hfill
    \caption{Qualitative comparison of snow removal between DSOR, DDIOR, and our DMNR. (a) Raw Point Cloud;\space(b) DSOR;\space(c) DDIOR;\space(d) DMNR. The left-hand image of each sub-image is the overall visualization of de-noising effect while the right-hand image is a zoomed-in version of that on the left. Colors of the point cloud are rendered according to labels, and noise points are those in \textcolor{yellow}{yellow}.}
    \label{fig:overall vis1}
\end{figure*}

We compare our methods with DSOR \cite{Kurup2021DSORAS} and DDIOR \cite{Wang2022ASA} on WADS \cite{Kurup2021DSORAS} and DENSE \cite{Bijelic_2020_STF} for both de-snowing and de-fogging. WADS are scanned by 64-channel LiDAR sensors under snowy weather and the snow particles are annotated manually. DENSE are scanned by 32-channel LiDAR sensors under clear weather. Snow simulation \cite{Hahner2022LiDARSS} with snowfall rate $r_s=2mm/h$ and fog simulation \cite{Hahner2021FogSO} with spatially varying attenuation coefficient $\alpha=0.015$ are employed to synthesis snow and fog data over 1,800 frames point clouds in DENSE.
Since not all sequences are available in WADS (some sequences are labeled incorrectly or have too few frames), we use sequences 13, 23-24, 26, 28, 30, and 34-36 for experiments and visualization. 
All the experiments are conducted on a laptop with an Intel~\circledR\space Core\textsuperscript{TM} i5-8265U CPU and 8GB RAM. 
We use the default parameters of DSOR~\cite{Kurup2021DSORAS} and DDIOR~\cite{Wang2022ASA}. 
Table~\ref{table:1} shows the parameters of DMNR and DMNR-H in all the experiments.
\begin{table*}[htbp]
    \begin{center}
    \caption{Parameters of DMNR and DMNR-H.}
    \label{table:1}
    \setlength{\tabcolsep}{8mm}{
    \begin{tabular}{c c c c c}
    \hline   \textbf{$K$} & \textbf{$k_1$} & \textbf{$k_2$} & $k_3$ & \textbf{$h$}\\
    \hline   10 & 0.015 & 0.055 & 100 & 5\\
    \hline
    \end{tabular}}
    \end{center}
\end{table*}

\begin{figure*}[!ht]
    \centering
    \subfigure[Raw Point Cloud]{
        \label{Raw reverse}
        \includegraphics[width=0.41\linewidth]{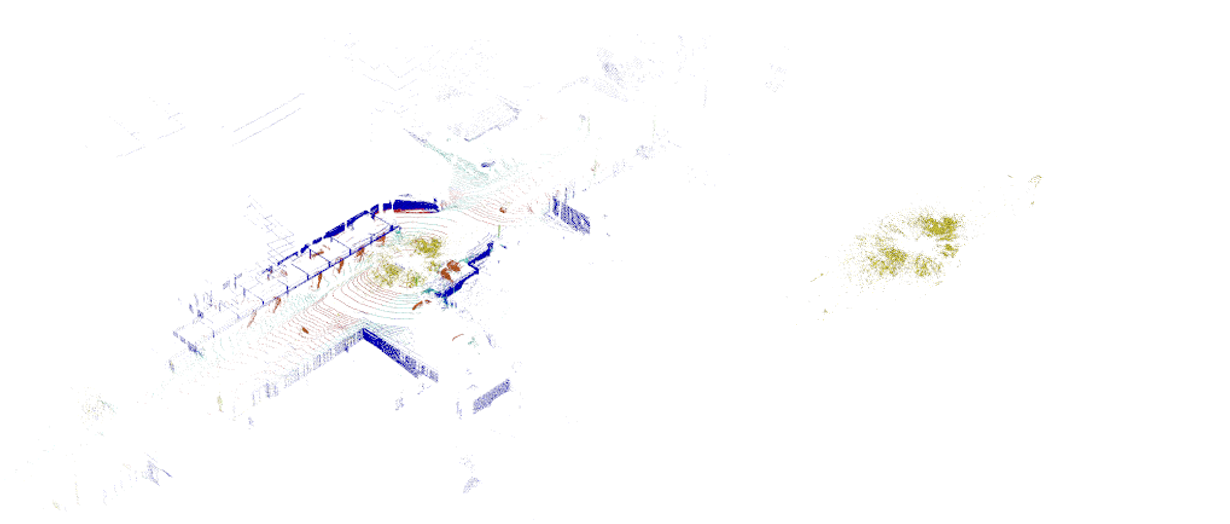}}
    \subfigure[DSOR]{
        \label{DSOR reverse}
        \includegraphics[width=0.41\linewidth]{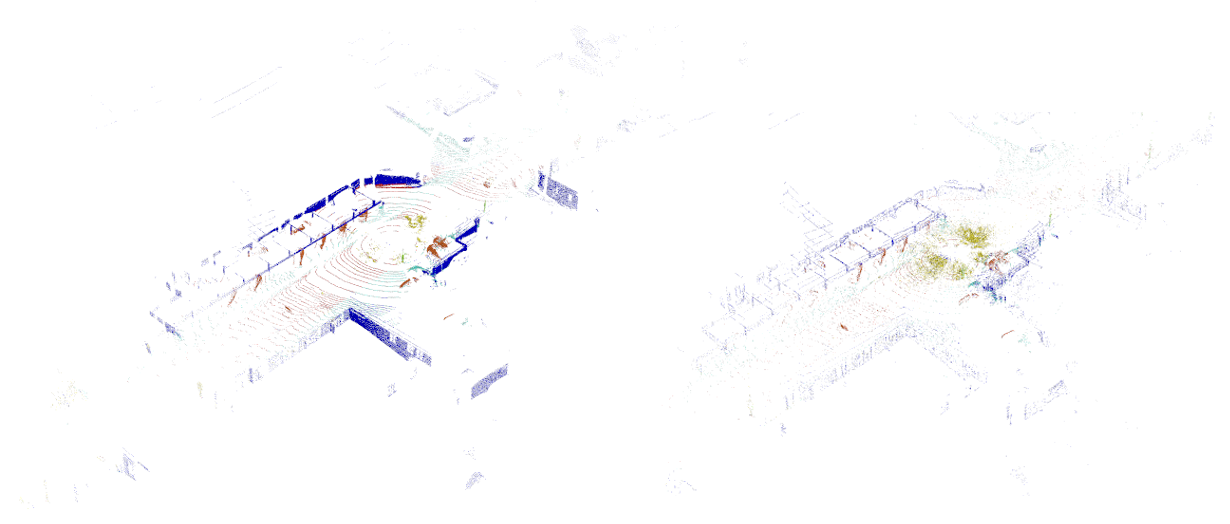}}
    \subfigure[DDIOR]{
        \label{DDIOR reverse}
        \includegraphics[width=0.41\linewidth]{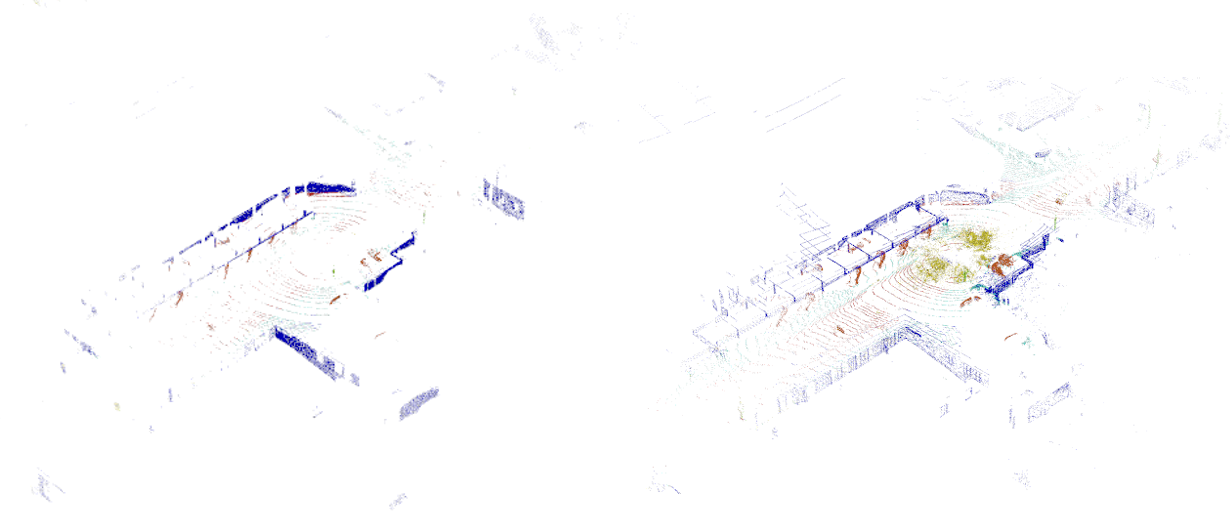}}
    \quad
    
    \subfigure[DMNR]{
        \label{DMNR reverse}
        \includegraphics[width=0.41\linewidth]{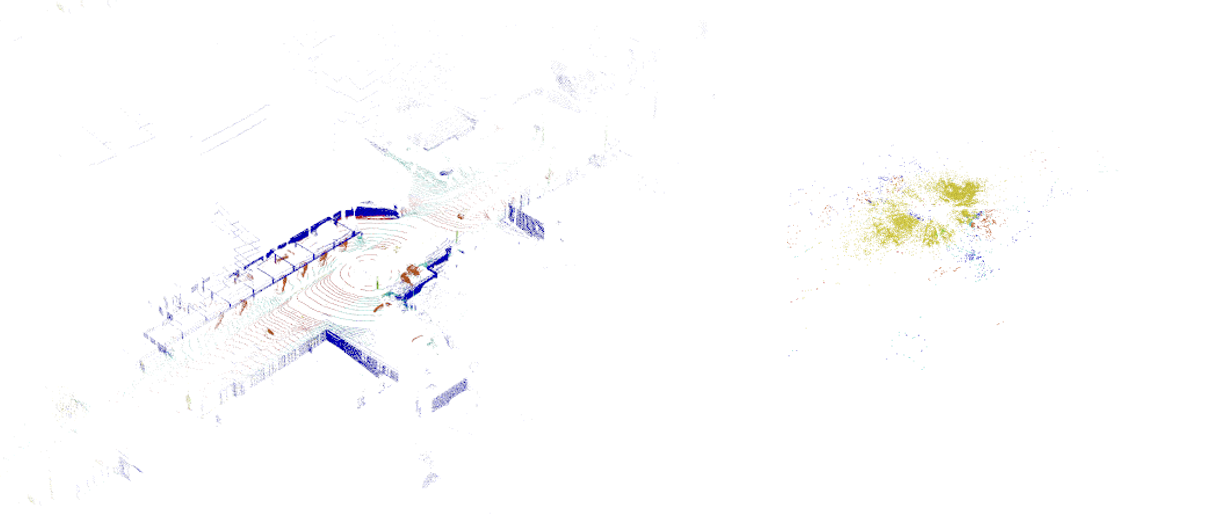}}
    \subfigure[DMNR-H]{
        \label{DMNR-H reverse}
        \includegraphics[width=0.41\linewidth]{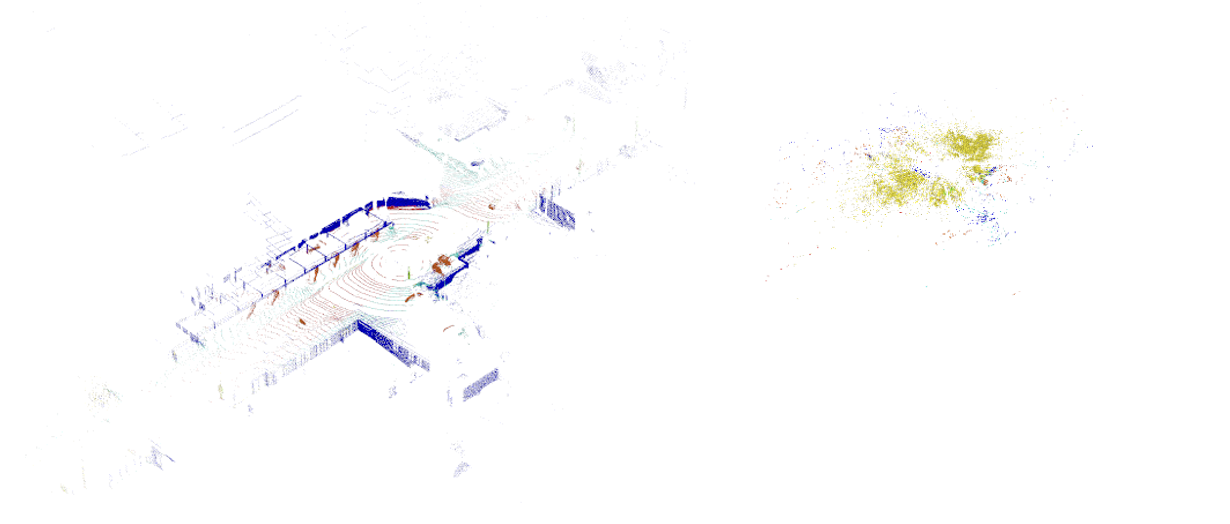}}
    \hfill
    
    \caption{Visualization of de-noising effect (left) and noise points judged by the corresponding filter (right). (a) Raw Point Cloud and Ground Truth Noise;\space(b) DSOR;\space(c) DDIOR;\space(d) DMNR;\space(d) DMNR-H. Colors of the point cloud are rendered according to labels, and noise points are those in \textcolor{yellow}{yellow}.}
    \label{fig:overall vis2}
\end{figure*}

\subsection{Qualitative Evaluation}
Fig.\ref{fig:preprocessing function} shows the effect of combining height threshold at pre-processing stage, where DMOR performs the best on preserving environmental details with high altitude. The overall de-noising effect is shown in Fig.\ref{fig:overall vis1} and Fig.\ref{fig:overall vis2} which visualized the point cloud that retained and removed respectively. By comparing with the raw point cloud image (Fig.\ref{Raw}) and referring to the color of the point cloud rendered by labels, DSOR removed large part of the environmental details (Fig.\ref{DSOR reverse}) while retained lots of noise  (Fig.\ref{DSOR}). Although DDIOR removed noise thoroughly (Fig.\ref{DDIOR}), even more environmental features were removed (Fig.\ref{DDIOR reverse}) than DSOR were. Our DMOR filter can remove noise completely (Fig.\ref{DMNR}) while nearly no environmental information was misjudged (Fig.\ref{DMNR reverse}).

% Fig.~\ref{fig:preprocessing function} shows the effect of combining height threshold at the environmental structure preservation stage, where DMNR performs the best on preserving environmental details with high altitude. 
Besides Fig.~\ref{fig:preprocessing function}, the overall de-noising effects are also shown in Fig.~\ref{fig:overall vis1} and Fig.~\ref{fig:overall vis2}. 
% The two figures show the point clouds that retained and removed respectively. 
By comparing with the raw point cloud (Fig.~\ref{Raw}), DSOR removes a large part of the environmental details (Fig.~\ref{DSOR reverse}) while retaining lots of noise points (Fig.~\ref{DSOR}). Although DDIOR removes noise more thoroughly (Fig.~\ref{DDIOR}), more environmental points are also removed (Fig.~\ref{DDIOR reverse}) than DSOR. Our DMNR and DMNR-H filter can remove noise more completely (Fig.~\ref{DMNR} and Fig.~\ref{DMNR-H reverse}) while nearly no environmental information is misjudged (Fig.~\ref{DMNR reverse} and Fig.~\ref{DMNR-H reverse}).

\begin{table}[htbp]
\begin{center}
\caption{Quantitative Comparison on de-snowing}
\label{table:2}
\begin{threeparttable}
\setlength{\tabcolsep}{0.5mm}{
\begin{tabular}{c c c c c c c}
\hline  \multirow{2}{*}{\textbf{Filters}}& & \textbf{WADS\cite{Kurup2021DSORAS}}\tnote{*} & & & \textbf{DENSE~\cite{Bijelic_2020_STF}}\tnote{**} & \\ & \textbf{$precision$} & \textbf{$recall$} & \textbf{$F1$} & \textbf{$precision$} & \textbf{$recall$} & \textbf{$F1$}\\
\hline   DSOR\cite{Kurup2021DSORAS} & 65.07 & 95.60 & 77.43 & 5.60 & 98.37 & 10.60 \\
   DDIOR\cite{Wang2022ASA} & 69.87 & 95.23 & 80.60 & 18.38 & 91.47 & 30.60 \\
   DMNR & 91.82 & 90.69 & 91.25 & 80.25 & 61.61 & 69.71 \\
   DMNR-H & 92.86 & 89.88 & \textbf{91.35} & 83.52 & 60.03 & \textbf{69.86} \\
   WeatherNet\tnote{$\dagger$}~\cite{Heinzler_2020} & 94.88 & 85.13 & 89.74 & - & - & - \\
\hline
\end{tabular}}
\begin{tablenotes}
\footnotesize
\item{*} 909 frames of outdoor collected snow. {**} 1,816 frames of simulated snow~\cite{Hahner2022LiDARSS}.
\item{$\dagger$} Deep learning based method.
\end{tablenotes}
\end{threeparttable}
\end{center}
\end{table}

\subsection{Quantitative Evaluation}
Generally, $precision$ and $recall$ are used for quantitative assessment of de-noising performance:
\begin{equation}
  precision = \frac {TP} {TP+FP} , recall = \frac {TP} {TP+FN}.
  \label{eq:precision&recall}
\end{equation}
$TP$ (True Positive) refers to noise points filtered out from the point cloud. $FP$ (False Positive) refers to non-noise points filtered out from the point cloud. $FN$ (False Negative) refers to noise points retained after filtering. 
As its definition, $precision$ evaluates the ability of a filter to maintain environmental details while $recall$ evaluates the ability to remove noise points. An aggressive filter increases $recall$ but decreases $precision$ while it would be an opposite effect for a mild filter, in which case we will not be able to make a comparison among filters. Thus, a comprehensive value for assessment is necessary. We use $F1-score$, denoted by $F1$, to comprehensively evaluate the de-noising ability of filters:
\begin{equation}
  F1 = \frac {2} {\frac{1}{recall}+\frac{1}{precision}}.
  \label{eq:F1 def}
\end{equation}

Table~\ref{table:2} shows the quantitative comparison among DSOR, DDIOR, DMNR, DMNR-H and WeatherNet~\cite{Heinzler_2020} on WADS~\cite{Kurup2021DSORAS} and DENSE~\cite{Bijelic_2020_STF} datasets.
The results on the WADS dataset illustrate that DSOR and DDIOR filters have higher $recall$ but lower $precision$. While our methods have the highest $F1$, $91.25$ and $91.35$, which are higher than the deep learning based supervised method WeatherNet (89.74), and outperform DSOR and DDIOR by a large margin (77.43 and 80.60).

\begin{table*}[htbp]
\begin{center}
\caption{Quantitative Comparison on de-fogging}
\label{table:3}
\begin{threeparttable}
\setlength{\tabcolsep}{6mm}{
\begin{tabular}{c c c c}
\hline  \multirow{2}{*}{\textbf{Filters}}& &  \textbf{fog}\tnote{*} \\ & \textbf{$precision$} & \textbf{$recall$} & \textbf{$F1$} \\
\hline   DSOR\cite{Kurup2021DSORAS} & 23.74 & 99.99 & 38.37 \\
   DDIOR\cite{Wang2022ASA} & 26.47 & 99.88 & 41.85 \\
   DMNR & 82.36 & 83.26 & 81.73 \\
   DMNR-H & 83.49 & 82.36 & \textbf{81.77} \\
\hline
\end{tabular}}
\begin{tablenotes}
\footnotesize
\item{*} 1,816 frames of simulated fog~\cite{Hahner2021FogSO} in DENSE~\cite{Bijelic_2020_STF}.
\end{tablenotes}
\end{threeparttable}
\end{center}
\end{table*}

The results of DSOR and DDIOR filters on the DENSE dataset show that they perform poorly for different kinds of snow and dog. 
The point clouds of DENSE dataset are scanned with 32-channel LiDARs, and have less number of points, smaller scanning range, and higher overall intensity than that of WADS dataset. Thus they own sparser density and a huge variety of intensity when corrupted by airborne particles. However, our methods still maintain higher $F1$ for the two cases. What's more, with the help of HDBSCAN, DMNR-H achieves even better $F1$ than DMNR for all the cases.

\section{Conclusions}
\label{sec:conc}
We propose an unsupervised de-noising algorithm DMNR for airborne particle removal in LiDAR data. It includes an environmental structure preservation stage and a dynamic filtering stage. The first stage effectively protects sparse environment points at high places. The second stage identifies noise points by an adaptive threshold considering density, intensity, and range together.
% and a dynamic threshold that is objectively consistent with the statistical distribution differences between noise and non-noise points.
Besides, DMNR-H is proposed by appending a HDBSCAN clustering algorithm to further improve DMNR's ability to preserve environmental information.

Thanks to the excellent ability of both environmental structure preserving and noise filtering, our methods far outperform the state-of-the-art unsupervised methods in terms of comprehensive de-noising ability when processing point clouds corrupted by various of airborne particles obtained by different LiDAR devices.

%Bibliography
\bibliographystyle{unsrt}  
\bibliography{references}

\end{document}